\pdfoutput=1

\documentclass[11pt]{article}

\usepackage[]{acl}

\usepackage{times}
\usepackage{latexsym}

\usepackage[T1]{fontenc}

\usepackage[utf8]{inputenc}

\usepackage{microtype}
\usepackage{comment}

\usepackage{xcolor}
\usepackage{booktabs}

\usepackage{multirow,array,float,graphicx}
\usepackage{amsmath,amssymb}
\usepackage{enumitem}

\usepackage[normalem]{ulem}
\useunder{\uline}{\ul}{}
\usepackage{setspace}

\newcommand{\jose}[1]{\textcolor{red}{#1}}

\newcommand{\fra}[1]{\textcolor{orange}{#1}}

\newcommand\blfootnote[1]{%
  \begingroup
  \renewcommand\thefootnote{}\footnote{#1}%
  \addtocounter{footnote}{-1}%
  \endgroup
}

\title{TempoWiC: An Evaluation Benchmark \\ for Detecting Meaning Shift in Social Media}

\author{\textbf{Daniel Loureiro$^{\diamondsuit}$} \\ \textbf{Aminette D'Souza$^{*\diamondsuit}$, Areej Nasser Muhajab$^{*\diamondsuit}$, Isabella A. White$^{*\diamondsuit}$, Gabriel Wong$^{*\diamondsuit}$}\\ \textbf{Luis Espinosa Anke$^{\diamondsuit\heartsuit}$, Leonardo Neves$^{\clubsuit}$, Francesco Barbieri$^{\clubsuit}$, Jose Camacho-Collados$^{\diamondsuit}$} \\
  $^{\diamondsuit}$ Cardiff NLP, School of Computer Science and Informatics, Cardiff University, UK  \\
 $^{\clubsuit}$ Snap Inc., Santa Monica, California, USA; $^{\heartsuit}$AMPLYFI, UK
  \\ 
   $^{\diamondsuit}$\texttt{boucanovaloureirod@cardiff.ac.uk, cardiffnlp.contact@gmail.com}\\
}

\begin{document}
\maketitle


\begin{abstract}
\vspace{-8pt} 
Language evolves over time, and word meaning changes accordingly. This is especially true in social media, since its dynamic nature leads to faster semantic shifts, making it challenging for NLP models to deal with new content and trends. However, the number of datasets and models that specifically address the dynamic nature of these social platforms is scarce. To bridge this gap, we present TempoWiC, a new benchmark especially aimed at accelerating research in social media-based meaning shift. Our results show that TempoWiC is a challenging benchmark, even for recently-released language models specialized in social media.
\blfootnote{$^{*}$Annotation team.}

\end{abstract}

\section{Introduction}
\vspace{-6pt} 

One of the most studied challenges in NLP is lexical ambiguity.
Solutions include word sense disambiguation \cite{navigli2009word} or entity linking \cite{ling2015design}, where words are linked to sense inventories such as WordNet \cite{miller1995wordnet} or Wikipedia. 
Recently, notable progress has been made with the advent of Language Models (LMs) and contextualized embeddings, crucially well equipped for modeling meaning in context \cite{pilehvar2020embeddings,bevilacqua2020generationary}.

One notable limitation with current lexical semantics benchmarks, however, is that they are typically clean and time-invariant, where standard grammar is the norm, and have little to no account of language usage in real-world platforms like social media.
%
However, there is ample agreement that modeling changes in language and topic distributions is crucial for modern NLP \cite{loureiro-etal-2022-timelms}.
Thus, there is a rich body of literature concerned with, e.g., adapting existing word representations (mainly word embeddings) diachronically \cite{hamilton-etal-2016-diachronic,szymanski-2017-temporal,rosenfeld-erk-2018-deep,hofmann-etal-2021-dynamic}, exposing LMs to time-specific data \cite{lazaridou2021pitfalls}, or temporal adaptation in general \cite{luu2021time,agarwal2021temporal,jin2021lifelong,loureiro-etal-2022-timelms}.
The lack of real-world data to serve as ground truth has typically limited the evaluation of diachronic word-level NLP models.
This limitation has been addressed in a myriad of ways, e.g., by comparing distributional similarities with human judgments \cite{gulordava2011distributional}, contrasting change vs. frequency \cite{hamilton-etal-2016-diachronic}, comparing time-sensitive representations with stripped-down versions \cite{frermann2016bayesian} or, more recently, determining whether a word has acquired new senses over time by looking at relatively large targeted subcorpora \cite{van2018semeval}, or probing for diachronic awareness in settings reminiscent of knowledge base completion \cite{dhingra2021time,hofmann-etal-2021-dynamic}.

\begin{table*}[ht]
\centering
\resizebox{\textwidth}{!}{%
\begin{tabular}{@{}ccc@{}}
\toprule
\textbf{Tweet 1} & \textbf{Tweet 2} & \textbf{Label} \\ \midrule


\textit{2019-02} & \textit{2020-02} & \\
"I ain't gone let the ppl \textit{frisk} me & "Set up a stop and \textit{frisk} outside a & T \\
if I'm dirty homie" &  white club and catch coke heads" & \\ \midrule

\textit{2019-04} & \textit{2020-04} & \\
"i wish i still had images of my old animal crossing new leaf \textit{villager} & "How does \textit{villager} trading in New horizons even work & T \\
 he was good boy" &   like tf" & \\ \midrule
 
\textit{2019-08} & \textit{2020-08} & \\
"This dude just said "Boys of the Backstreet" He made em sound & "my target app said they didn't have \textit{folklore} cds but & F \\
 like a whole \textit{folklore}" &   when i went inside they had some i'm so happy" & \\ \midrule 
 
\textit{2019-08} & \textit{2020-08} & \\
"In case you were wondering facial devotion & "With these \textit{mask} at work customers are & F \\
still worked with a face \textit{mask} on" &   forever confusing me and Reyna lmao" & \\

\bottomrule
\end{tabular}%
}
\caption{Examples from the training set of TempoWiC. Target words in \textit{italic}. The label T (True) indicates that the word has the same meaning in the two tweets, the label F (False) indicates that the meaning is different.
}
\label{tab:examples}
\end{table*}

Despite the above, few works have attempted to model the connection between meaning shift and social media.
Among them, \newcite{del-tredici-etal-2019-short} showed that trending words are a meaningful signal for predicting meaning shifts. This platform-specific insight is the main basis for the construction of our dataset. TempoWiC follows the simple formulation from the SuperGLUE Word-in-Context (WiC) challenge \cite{pilehvar-camacho-collados-2019-wic}, which is particularly well suited for temporal meaning shift evaluation given that it is not reliant on a reference sense inventory.
This change of paradigm has seen wide adoption, with multilingual extensions such as XL-WiC \cite{raganato-etal-2020-xl}, Am2ico \cite{liu-etal-2021-am2ico} or MCL-WiC \cite{martelli-etal-2021-semeval}, or reformulations such as WiC-TSV \cite{breit-etal-2021-wic}.
In contrast to these, TempoWiC is crucially designed around meaning shift and instances of word usage tied to Twitter trending topics.
As an additional contribution, along with the benchmark, we provide a set of robust baselines and analyses that highlight the challenging nature of the task.

\section{TempoWiC: Temporal Word in Context}
\label{dataset}

In this section, we describe our process to build our evaluation benchmark for detecting meaning shift in social media. 
The task is framed as a simple binary classification problem in which a target word is present in two texts (tweets) posted during different time periods. The goal is to decide whether the meaning corresponding to the first target word in context is the same as the second one or not. Table \ref{tab:examples} lists a few examples. 

\subsection{Data collection}
\label{datacollection}

\paragraph{Word Selection.}
Since this work focuses on meaning shift, we do not consider neologisms and use lemmas from WordNet as an initial set of potential words of interest (82K lemmas, ignoring multi-word expressions, stopwords and numbers). 
From a corpus of 100M tweets collected from the Twitter API for the period between the start of 2019 and September of 2021, we compiled monthly frequency counts for this set of known words, and computed trending scores following \citet{chen-etal-2021-mitigating}.
Each trending word peak is estimated as the day with highest frequency during the year/month with most occurrences.
As the prior date, we considered the same date exactly one year before.
This is done in order to avoid seasonal confound factors, which are known to affect models in social media \cite{chae2012spatiotemporal,barbieri2018exploring}.
Afterwards, we selected the top 10 words with highest trending scores from each month, resulting in 210 words which are candidates for annotation. For this selection, we ignored words with fewer than 100 occurrences in our corpus during their peak date.

\paragraph{Obtaining Paired Tweets.}

In this phase we collected, for each trending word, 100 tweets posted during the peak date, and 100 tweets posted during the prior date.
For this phase we used the Twitter APIs, setting filters to request only English tweets, and ignoring replies and retweets.\footnote{We retrieved additional tweets with the APIs as we included trending words that had a minimum occurrences of 100 tweets on peak day in our initial dataset, but some words included less than 100 tweets on the prior date.}
We preprocessed each tweet using spaCy \cite{Honnibal_spaCy_Industrial-strength_Natural_2020} and we randomly paired tweets from the prior and peak sets for specific words that match both in surface form and part-of-speech tag.

\subsection{Annotation}
\label{annotation}

\paragraph{Annotators.} We recruited four annotators through our internal institution recruitment office. This ensured that the annotators were part of the process, trained and understood all the details of the task.\footnote{The full guidelines provided to the annotators are available in the task website.} Annotators, who were all native or near-native English speakers, were all paid the equivalent of a research assistant per hour.

\paragraph{First stage.} The annotation was split into two phases. First, we took a breadth-first approach in which a relatively short number of instances (i.e., 10) of a large number of the selected words (210 in total) were annotated. The motivation for this initial phase was to understand which words had some sort of meaning shift to start with. The selection of the words to be included in the dataset was then restricted to words which had more than 3 out of 10 instances with meaning shift.

\paragraph{Second stage.} The second phase was based on a depth-first approach in which 100 instances of all selected words from the first stage were annotated. We ensured that each instance was annotated by three annotators. The final label attributed to each instance was determined by majority vote.\footnote{Some words proved too difficult to reliably annotate according to feedback from the annotators as well as low agreement scores computed after annotation (more details in Section \ref{stats}). Consequently, these words were removed from the dataset. Among the various challenges to be expected from annotating social media, mixed language (e.g., English and Hindi) was among the most frequent issues.}

\begin{table}[t]
\centering
\resizebox{\columnwidth}{!}{%
\begin{tabular}{@{}clccc@{}}
\toprule
\multicolumn{1}{l}{\multirow{2}{*}{\textbf{}}} & \multicolumn{1}{c}{\multirow{2}{*}{\textbf{Word}}} & \textbf{\# Instances} & \textbf{Trending} & \textbf{Agreement} \\ 
\multicolumn{1}{l}{} & \multicolumn{1}{c}{} & \textbf{\small{(\% Diff. Meaning)}} & \textbf{Date} & \textbf{\small{(Krippendorf's $\alpha$)}} \\ \midrule \midrule
\multirow{15}{*}{\rotatebox{90}{\textbf{Train}}} & frisk & 99 (54\%) & 11/2/2020 & 0.718 \\
 & pogrom & 99 (5\%) & 25/2/2020 & 0.482 \\
 & containment & 100 (33\%) & 12/3/2020 & 0.274 \\
 & virus & 96 (48\%) & 12/3/2020 & 0.254 \\
 & epicenter & 100 (71\%) & 14/3/2020 & 0.124 \\
 & ventilator & 99 (17\%) & 27/3/2020 & 0.541 \\
 & villager & 100 (64\%) & 10/4/2020 & 0.546 \\
 & turnip & 100 (95\%) & 10/5/2020 & 0.316 \\
 & bunker & 98 (61\%) & 1/6/2020 & 0.408 \\
 & mask & 99 (76\%) & 14/7/2020 & 0.255 \\
 & teargas & 98 (3\%) & 18/7/2020 & 0.786 \\
 & paternity & 100 (22\%) & 30/7/2020 & 0.289 \\
 & entanglement & 99 (89\%) & 1/8/2020 & 0.623 \\
 & folklore & 82 (92\%) & 3/8/2020 & 0.917 \\
 & parasol & 100 (85\%) & 2/9/2020 & 0.446 \\ \cmidrule(lr){1-5}
\multirow{4}{*}{\rotatebox{90}{\textbf{Validation}}} & impostor & 99 (76\%) & 23/9/2020 & 0.544 \\
 & lotte & 98 (43\%) & 27/9/2020 & 0.514 \\
 & recount & 100 (28\%) & 6/11/2020 & 0.682 \\
 & primo & 100 (77\%) & 9/11/2020 & 0.528 \\   \cmidrule(lr){1-5}
\multirow{15}{*}{\rotatebox{90}{\textbf{Test}}} & milker & 99 (50\%) & 4/3/2021 & 0.699 \\
 & moxie & 97 (83\%) & 5/3/2021 & 0.755 \\
 & unlabeled & 100 (90\%) & 10/3/2021 & 0.711 \\
 & pyre & 100 (32\%) & 27/4/2021 & 0.243 \\
 & gaza & 100 (60\%) & 15/5/2021 & 0.749 \\
 & ido & 91 (83\%) & 27/5/2021 & 0.712 \\
 & airdrop & 99 (40\%) & 6/6/2021 & 0.918 \\
 & bullpen & 99 (9\%) & 16/6/2021 & 0.388 \\
 & crt & 100 (68\%) & 26/6/2021 & 0.867 \\
 & monet & 98 (94\%) & 8/7/2021 & 1.000 \\
 & burnham & 100 (16\%) & 1/8/2021 & 0.964 \\
 & delta & 100 (100\%) & 11/8/2021 & 1.000 \\
 & gala & 100 (46\%) & 14/9/2021 & 0.498 \\
 & launchpad & 99 (81\%) & 17/9/2021 & 0.558 \\
 & vanguard & 99 (95\%) & 21/9/2021 & 0.421 \\ \bottomrule
\end{tabular}%
}
\caption{Details all the words included in TempoWiC. Maximum pairwise agreement is reported.}
\label{tab:word_details}
\end{table}

\subsection{Statistics and Inter-annotator Agreement}
\label{stats}

The outcome of our annotation pipeline is a dataset of 3,297 instances divided in train/validation/test sets of size 1,428/396/1,473 instances, respectively.
We measured inter-annotator agreement using Fleiss' Kappa at 0.446, and using Krippendorff's $\alpha$ at 0.439.
Since each instance is assigned a majority vote label, we also computed the maximum pairwise Krippendorff's $\alpha$ at 0.627, which should be more revealing of the expected performance on this task. Words with Krippendorff's $\alpha$ below 0.1 were removed from the dataset. Table \ref{tab:word_details} provides a summary of the most relevant details of the dataset after annotation.

\section{Evaluation}

In this section, we report baseline results on TempoWiC using two different approaches which have proven successful on the WiC task that inspired this work.
More concretely, we report results based on pretrained LMs using fine-tuning on the tweet pair as well as comparing the similarity of contextual embeddings.\footnote{Additional baselines are reported in Appendix \ref{sec:appendix_more_baselines}.}

\paragraph{Evaluation metrics.} The results are reported according to the standard Macro-F1 metric for multi-class classification problems. Accuracy is also reported for completeness but, given the unbalanced nature of the dataset, Macro-F1 should provide a more accurate representation of the performance.

\subsection{Models}

Our experiments include the following LMs: RoBERTa base and large pretrained on general domain corpora \cite{roberta}; RoBERTa base with continued training on tweets until the end of 2019, and a similar model trained with more tweets until the end of 2021 \cite[TimeLMs]{loureiro-etal-2022-timelms}; LMs based on RoBERTa but trained from scratch on tweets, both base and large versions \cite[BERTweet]{nguyen-etal-2020-bertweet}.

Each of these LMs is fine-tuned representing instances as "\texttt{<s>} Tweet 1 \texttt{</s>} Tweet 2 \texttt{</s>}", with each tweet represented by the encodings produced by each model's tokenizer.\footnote{We experimented concatenating the encodings for the target word at the end of the sequence as proposed by \citet{NEURIPS2019_4496bf24} for WiC, but found no improvements.}
Additionally, we trained a logistic regression classifier on the cosine similarity of the contextual embeddings corresponding to the target word on each tweet of the pair.
This approach based on contextual embeddings is sensitive to the choice of layers from the LM used to represent embeddings.
Following \citet{LOUREIRO2022103661},
we use SP-WSD layer pooling weights (model specific) that are suited for sense representation (see Appendix \ref{sec:appendix_pooling} for ablation results with alternative pooling strategies). Both approaches are implemented with \citet[Transformers]{wolf-etal-2020-transformers}.

\subsection{Results}

\begin{table}[t]
\centering
\resizebox{\columnwidth}{!}{%
\begin{tabular}{@{}clcc@{}}
\toprule
\multicolumn{1}{l}{} & \multicolumn{1}{c}{\textbf{Model}} & \textbf{Accuracy} & \textbf{Macro-F1} \\ \midrule
\multirow{6}{*}{\rotatebox{90}{Fine-tuning}} & RoBERTa-base & 66.89\% & 58.26\% \\
 & RoBERTa-large & 66.49\% & 59.10\% \\
 & TimeLMs-2019-90M & 66.46\% & 57.70\% \\
 & TimeLMs-2021-124M & 65.04\% & 54.75\% \\
 & BERTweet-base & 61.46\% & 51.27\% \\
 & BERTweet-large & {\ul 67.93\%} & {\ul 60.62\%} \\ \midrule
\multirow{6}{*}{\rotatebox{90}{Similarity}} & RoBERTa-base & 67.96\% & 52.89\% \\
 & RoBERTa-large & 72.98\% & 67.09\% \\
 & TimeLMs-2019-90M & {\ul \textbf{74.07\%}} & {\ul \textbf{70.33\%}} \\
 & TimeLMs-2021-124M & 71.01\% & 63.51\% \\
 & BERTweet-base & 69.45\% & 65.16\% \\
 & BERTweet-large & 69.18\% & 56.95\% \\ \midrule
\multirow{3}{*}{\rotatebox{90}{Naive}} & Random & 50.00\% & {\ul 50.00\%} \\
 & All True & 36.59\% & 26.79\% \\
 & All False & {\ul 63.41\%} & 38.80\% \\
 \bottomrule
\end{tabular}%
}
\caption{Main results on the test set of TempoWiC. Fine-tuning results are the average of 3 runs.}
\label{tab:main_results}
\end{table}

Our results on Table \ref{tab:main_results} show that TempoWiC is a challenging task with room for improvement.
While the best results using both fine-tuning and similarity approaches are obtained by models adapted to the Twitter domain, this advantage isn't substantial over generic RoBERTa. Interestingly, we find that the straightforward similarity approach manages to substantially outperform fine-tuning, with a Twitter base model trained with data before any word's trending peak achieving the best performance.
While this result may be surprising considering that fine-tuning performs better on WiC, this finding is in line with recent work in word sense disambiguation showing that approaches based on contextual embeddings can be more robust and generalizable than fine-tuning \cite{10.1162/coli_a_00405}.

\begin{table}[t]
\centering
\resizebox{\columnwidth}{!}{%
\begin{tabular}{@{}l|cc|cc@{}}
\toprule
 \multirow{2}{*}{\textbf{Word}} & \multicolumn{2}{c|}{\textbf{Accuracy}} & \multicolumn{2}{c}{\textbf{Macro-F1}} \\
 & \textbf{Fine-tune} & \textbf{Similarity} & \textbf{Fine-tune} & \textbf{Similarity} \\ \midrule
 airdrop & 40.48\% & \textbf{65.31\%} & 30.13\% & \textbf{65.18\%} \\
 bullpen & \textbf{67.68\%} & 38.38\% & \textbf{44.53\%} & 34.54\% \\
 burnham & 27.67\% & \textbf{83.00\%} & 27.62\% & \textbf{69.15\%} \\
 crt & 64.98\% & \textbf{79.80\%} & 46.26\% & \textbf{73.24\%} \\
 delta & 85.37\% & \textbf{98.98\%} & 46.05\% & \textbf{49.74\%} \\
 gala & 47.67\% & \textbf{78.00\%} & 36.71\% & \textbf{77.86\%} \\
 gaza & 67.01\% & \textbf{69.39\%} & \textbf{66.74\%} & 64.61\% \\
 ido & 83.88\% & \textbf{90.11\%} & 63.04\% & \textbf{75.78\%} \\
 launchpad & \textbf{81.82\%} & 80.81\% & 45.00\% & \textbf{59.26\%} \\
 milker & 46.13\% & \textbf{64.65\%} & 37.28\% & \textbf{63.44\%} \\
 monet & 93.54\% & \textbf{94.90\%} & 48.33\% & \textbf{62.96\%} \\
 moxie & \textbf{89.35\%} & 68.04\% & \textbf{74.75\%} & 56.48\% \\
 pyre & \textbf{64.29\%} & 51.02\% & \textbf{62.78\%} & 50.84\% \\
 unlabeled & 65.67\% & \textbf{76.00\%} & 47.71\% & \textbf{57.48\%} \\
 vanguard & \textbf{95.96\%} & 73.74\% & \textbf{68.80\%} & 42.44\% \\
 \bottomrule
\end{tabular}%
}
\caption{Performance by word on the TempoWiC test set. Using TimeLMs-2019-90M as the best similarity model, and BERTweet-large as the best fine-tuning model (average of 3 runs).}
\label{tab:word_results}
\end{table}

\paragraph{Analysis by word.} Table \ref{tab:word_results} provides a detailed breakdown of the results of the best performing model (i.e., TimeLMs-2019-90M) by individual words. As can be observed, there are large differences between words, which are also due to the unbalanced natural distribution of certain words to start with (see Table \ref{tab:word_details}). More interesting is perhaps the gaps between fine-tuning and similarity techniques. While similarity appears to be generally more robust, in words such as \textit{bullpen} or \textit{vanguard}, the tendency is reversed.

\subsection{Future Work}

This work only covers English, but future work should include additional languages and experiment with both multilingual and monolingual models.
We leave an analysis explaining the difference between the 2019 and 2021 models for future work as well, alongside the development of methods that leverage the dates provided with each instance towards improved performance, similarly to \citet{dhingra2021time, 10.1145/3488560.3498529}.

\section{Conclusion}

This work introduced a new lexical semantics task and a dataset, TempoWiC, focused on meaning shift detection in Twitter. 
While meaning representation is at the core of the task, the challenges of this task go beyond simple word sense disambiguation with a focus on its temporal aspect.
To make the task realistic, we extracted Twitter trending words for different periods and paired them with tweets from past periods.
This makes the task more challenging and grounded in real-world applications for social media.
We performed extensive experiments with standard meaning representation approaches based on language models.
The results show that the task leaves ample room for improvement, with several avenues for future research on how to better integrate time-aware social media models with meaning representation techniques.
The TempoWiC dataset and baseline scripts are available at \href{https://github.com/cardiffnlp/TempoWiC}{github.com/cardiffnlp/TempoWiC}.

\section*{Acknowledgements}

Jose Camacho-Collados and Daniel Loureiro are supported by a UKRI Future Leaders Fellowship.
We thank Snap Inc. for providing resources to support this project.


\bibliography{anthology,custom}
\bibliographystyle{acl_natbib}

\appendix

\clearpage

\section{Additional Baselines}
\label{sec:appendix_more_baselines}

Besides the fine-tuning and similarity methods described in the main paper, we also experimented with additional approaches in order to better understand the difficulty of this dataset.

In this appendix we provide additional results based on an MLP trained with concatenated contextual embeddings (Table \ref{tab:contextual_mlp}), and another MLP trained with the concatenation of the average of static embeddings from each tweet (Table \ref{tab:static_results}). Embeddings are \textit{L2} normalized after concatenation.

The hyper-parameters used with these MLPs were determined by grid search on 24 different configurations which were tested on the validation set.
The parameters tested were \textit{hidden layer sizes} ((\textit{embedding size}, 100) or (100)), \textit{solver} (adam or sgd), \textit{batch size} (32 or 64), and \textit{maximum number of iterations} (50, 100 or 200).

Static embeddings are based on fastText \cite{bojanowski-etal-2017-enriching} and learned from Twitter data on the same corpora used for \citet{loureiro-etal-2022-timelms}.
These embeddings are trained with skipgram for 300-dimensions, min-ngram size 2 and max-ngram size 12.

\begin{table}[ht]
\centering
\resizebox{\columnwidth}{!}{%
\begin{tabular}{@{}lcc@{}}
\toprule
\textbf{Model} & \textbf{Accuracy} & \textbf{Macro-F1} \\ \midrule
RoBERTa-base & 68.11\% (0.77\%) & 55.00\% (1.88\%) \\
RoBERTa-large & \textbf{68.82\%} (1.06\%) & 55.17\% (2.60\%) \\
TimeLMs-2019-90M & 68.68\% (0.68\%) & \textbf{58.62\%} (0.81\%) \\
TimeLMs-2021-124M & 68.23\% (0.90\%) & 57.55\% (3.57\%) \\
BERTweet-base & 65.51\% (0.91\%) & 57.03\% (3.56\%) \\
BERTweet-large & 66.55\% (0.59\%) & 54.16\% (0.09\%) \\
\bottomrule
\end{tabular}%
}
\caption{Performance of MLP trained with concatenation of the target word's contextual embeddings (SP-WSD pooling), tuned on the validation set. Reporting average of 3 runs, and standard deviation.}
\label{tab:contextual_mlp}
\end{table}

\begin{table}[ht]
\centering
\resizebox{\columnwidth}{!}{%
\begin{tabular}{@{}lcc@{}}
\toprule
\textbf{Model} & \textbf{Accuracy} & \textbf{Macro-F1} \\ \midrule
CommonCrawl & 55.53\% (0.78\%) & 48.98\% (0.57\%) \\
TimeLMs-2019-90M & \textbf{57.64\%} (0.71\%) & 49.46\% (0.28\%) \\
TimeLMs-2021-124M & 55.20\% (0.31\%) & \textbf{52.30\%} (0.57\%) \\
\bottomrule
\end{tabular}%
}
\caption{Performance of MLP trained with concatenation of the average of static embeddings from each tweet. Reporting average of 3 runs, and standard deviation.}
\label{tab:static_results}
\end{table}

\newpage

\section{Pooling Contextual Embeddings}
\label{sec:appendix_pooling}

Table \ref{tab:contextual_ablation} reports results using the similarity method described in the main paper with alternative choices for layer pooling.
We considered the final layer and the sum of the last 4 layers as these are common choices in word sense disambiguation settings.

\begin{table}[ht]
\centering
\resizebox{\columnwidth}{!}{
\begin{tabular}{@{}lccc@{}}
\toprule
\multicolumn{1}{l}{\textbf{Model}} & \textbf{Final Layer} & \textbf{Sum Last 4} & \textbf{SP-WSD} \\ \midrule
RoBERTa-base & 40.89\% & \textbf{60.33\%} & 52.89\% \\
RoBERTa-large & 38.80\% & 53.32\% & \textbf{67.09\%} \\
TimeLMs-2019-90M & 59.93\% & 67.69\% & \textbf{70.33\%} \\
TimeLMs-2021-124M & 53.14\% & 60.26\% & \textbf{63.51\%} \\
BERTweet-base & 67.35\% & \textbf{66.91\%} & 65.16\% \\
BERTweet-large & 38.80\% & 41.75\% & \textbf{56.95\%} \\
\bottomrule
\end{tabular}
}
\caption{Performance of Contextual Similarity method according to choice of layer pooling approach.}
\label{tab:contextual_ablation}
\end{table}

\end{document}